# Tri-Reader: An Open-Access, Multi-Stage AI Pipeline for First-Pass Lung Nodule Annotation in Screening CT

## Technical Development

**Fakrul Islam Tushar, PhD, Joseph Y. Lo PhD**

Center for Virtual Imaging Trials, Carl E. Ravin Advanced Imaging Laboratories, Department of Radiology, Duke University School of Medicine, Durham, NC

Dept. of Electrical & Computer Engineering, Pratt School of Engineering, Duke University, Durham

## 1. Introduction

Creating high-quality CT nodule annotations is time-consuming and expensive, limiting the scale and diversity of training data available for lung cancer screening AI. While datasets are increasingly becoming open-access and many are being curated through AI-human collaboration, these curation processes often remain private or commercial [1-4]. In recent years, the community has released multiple open-source implementations for lung segmentation [5, 6], nodule detection (computer-aided detection; *CADe*) [7-9], and malignancy classification (computer-aided diagnosis; *CADx*) [7, 10-13]. However, these tools are often used in isolation, and there is a practical need for an integrated, reproducible "first pass" pipeline that rapidly proposes a manageable set of candidates for downstream human review and annotation. Using multiple open-access models trained on public datasets, we developed Tri-Reader, a comprehensive, freely available pipeline that integrates lung segmentation, nodule detection, and malignancy classification into a unified tri-stage workflow [2, 3, 5, 7, 14, 15]. The pipeline is designed to prioritize sensitivity while reducing the candidate burden for annotators. To ensure accuracy and generalizability across diverse practices, we evaluated Tri-Reader on multiple internal and external datasets as compared with expert annotations and dataset-provided reference standards [3, 4, 16].

## 2. Materials and Methods
### 2.1. Datasets

We developed Tri-Reader using U.S.-based development cohorts spanning multiple institutions and acquisition eras [2, 3, 7, 14], and evaluated it on independent multinational test datasets from Europe (Portugal) [16] and Asia (China) [4] to assess generalizability across diverse clinical settings and reference standards. An overview of all datasets, including cohort size, age distribution, and annotation sources, is presented in **Table 1**. Because annotation conventions and ground-truth definitions vary across cohorts, detailed cohort descriptions and stratified analyses were reported in **Supplementary Tables S1.1, S1.2, S4.1**, **S4.5**, and **S4.14**, and in **Supplementary Sections S1.1, S4**, and **S4.3**.

### 2.2. Tri-Reader Pipeline

Tri-Reader consists of three sequential stages (**Figure 1**). **Stage 1** performs consensus *CADe* using two complementary detection models: **CADe-ROIonly** and **CADe-FPaware**, a false-positive-aware variant trained via strategic hard-negative mining. Both detection models are based on MONAI 3D RetinaNet trained on the Duke Lungs Cancer Screening (DLCS24) [3, 7], LUNA16 [14], LUNA25[2] and VLST [15] datasets. Candidates outside the lungs are rejected using lung masks produced by VISTA3D [5]. Candidates detected by both *CADe* variants are promoted with confidence 1.0 and passed to the final output. **Stage 2** applies ensemble malignancy scoring to disagreement candidates using two *CADx* classifiers trained on complementary data distributions: LUNA25-CADxr50 trained on LUNA25 and DLCS24-CADxr50SWS trained on DLCS with Strategic Warm-Start initialization [7]. For each candidate, the two models' malignancy scores were averaged. Candidates with averaged $CADx \geq \tau_{CADx}$ are promoted (confidence 0.5). **Stage 3** applies final *CADe* refinement: remaining candidates are retained if their averaged *CADe* score $\geq \tau_{CADe}$ (confidence 0.2). Operating thresholds (for both *CADe* and *CADx*) were determined through systematic evaluation on validation cohorts from DLCS, LUNA16, LUNA25, and VLST (**Supplement Section S1.4**), balancing sensitivity preservation with candidate burden reduction. The output is a single candidate list with confidence tiers (1.0, 0.5, 0.2) that can be used to seed annotation workflows.

When radiology reports are available, a rule-based NLP module can extract nodule descriptors (eg, size, lobe, laterality, Lung-RADS category) and automatically match them to Tri-Reader candidates based on spatial location and descriptor consistency. We include this as an extensible module to enable semi-automated label enrichment and retrospective cohort characterization.

### 2.3. Evaluation

We evaluated Tri-Reader on four datasets with diverse reference standards (**Table 1**): two internal screening cohorts (DLCS24 Test, n=198 scans, 294 nodules; DLCS Secret Test, n=419 scans, 651 nodules), the public LNDbv4 dataset with multi-radiologist consensus annotations and semantic characteristics (n=155 scans, 377 nodules), and the external IMD-CT cohort enriched for indeterminate nodules (n=2,032 scans, 2,032 nodules). This diversity enables assessment across different annotation protocols, nodule size distributions, and clinical contexts.

**For comparison**, we evaluated one baseline detectors, LUNA16-De, a MONAI RetinaNet model trained on LUNA16 following the standard 10-fold cross-validation protocol (fold 6 used for external validation) [3, 7]. This model was selected because it has been widely used in prior studies and in lung cancer screening pipelines to generate pseudo-labels for modern screening datasets curation [3, 7, 8, 14]. Both models use identical architectures and preprocessing (resampling to 0.7×0.7×1.25 mm, HU clipping -1000 to 500, patch size 192×192×80).

Evaluation was performed at the lesion level using free-response operating characteristic (FROC) analysis, reporting average sensitivity (CPM: mean sensitivity across 1/8, 1/4, 1/2, 1, 2, 4, and 8 false positives per scan) and candidates per scan [7, 14]. Beyond overall performance, we report dataset-specific sub-analyses: radiologist consensus patterns and semantic detectability factors (LNDbv4), size-stratified performance and benign/malignant subsets (IMD-CT), and qualitative review of missed cases.

## 3. Results

Expanded dataset-specific analyses (radiologist-consensus patterns, size- and diagnosis-stratified performance, candidate accounting, and failure case summaries) are provided in the Supplementary Material (**Tables S4.1-S4.15**).

### 3.1. Overall Performance

Table 2 summarizes lesion-level detection performance across internal and external test cohorts**.** On the DLCS benchmark test set, Tri-Reader achieved a CPM of 0.62 (95% CI, 0.58-0.67) and sensitivity of

0.68 at 1 false-positive per scan with 13.0 candidates per scan, compared with a CPM of 0.57 and 23.1 candidates per scan for LUNA16-De. On the DLCS private test set, Tri-Reader maintained improved efficiency with 12.4 candidates per scan and sensitivity of 0.60, versus 22.3 candidates per scan for the baseline. On the external IMD-CT cohort, Tri-Reader improved CPM from 0.63 to 0.73 and reduced candidates per scan from 13.9 to 6.71. On LNDbv4, Tri-Reader reduced candidates per scan from 19.3 to 11.34 while maintaining comparable CPM. Across all cohorts, Tri-Reader achieved a 40-55% reduction in candidate burden per scan relative to the baseline detector while preserving sensitivity.

### 3.2. Dataset-Specific Analyses

**LNDbv4 Consensus and Semantic Analysis.** Performance varied systematically with radiologist consensus patterns (**Supplementary Section S4.3.2, Table S4.11**). For three-reader unanimous consensus nodules (n=32), Tri-Reader assigned mean detection probability $0.94 \pm 0.13$, whereas single-reader nodules showed greater dispersion ($0.77 \pm 0.32$). Subtlety was the primary driver of detectability (Cohen's $d = 1.00$), with detected nodules significantly larger (6.47 mm vs 4.11 mm; $P < .001$) and more conspicuous than missed nodules. Tri-Reader demonstrated higher sensitivity to suspicious morphological features (Spiculation: $d = 0.51$; Malignancy rating: $d = 0.71$) compared with baseline model.

**IMD-CT Size Stratification and Malignancy Analysis.** Size-stratified analysis revealed expected performance scaling with nodule diameter (**Supplementary Table S4.15**). For clinically actionable nodules (10-20 mm; n=771), Tri-Reader achieved average sensitivity 0.76 with only 4.89 candidates/scan. Stage-wise composition analysis confirmed alignment with clinical priority: malignant prevalence was 81.2% in Stage 1 (high-confidence consensus), 68.5% in Stage 2 (CADx-promoted), and 66.7% in Stage 3 (CADe-refined). Error analysis showed missed nodules were predominantly sub-centimeter (median 5.90 mm for Tri-Reader unique misses) and disproportionately benign (15.2% miss rate for benign vs 5.8% for malignant).

**DLCS Pathological Correlation.** On DLCS24 test set, Tri-Reader identified 90.9% (30/33) of pathologically confirmed malignancies with median detection probability 0.99, demonstrating high confidence for cancer cases. The system identified 100% of Lung-RADS 4B nodules and 97.3% of nodules ≥10 mm (**Supplementary Section, S4.1.2 S4.2.2, Tables S4.3, S4.7**).

### 4. Discussion

Tri-Reader demonstrates that combining open-access implementations in a staged workflow can substantially reduce annotation workload without requiring new model training for each dataset. By integrating consensus detection, malignancy-informed recovery, and final refinement, the pipeline achieves ~2× candidate reduction while preserving sensitivity for clinically relevant nodules. This makes Tri-Reader well-suited as an initial pass for creating or expanding lung screening annotations, for active-learning review, or for curating candidate pools for consensus reading.

Importantly, the external datasets evaluated here use different reference standards (multi-reader consensus vs single-reader labels vs cohort-specific inclusion criteria) [4, 16]. Rather than attempting to harmonize these standards post hoc, we report both overall metrics and dataset-tailored analyses to clarify how performance changes with nodule size, reader agreement, and detectability factors [16]. On LNDbv4, the strong correlation between model confidence and radiologist consensus suggests Tri-Reader's internal scoring can serve as a proxy for annotation priority. The optional report linkage module extends Tri-Reader's utility by enabling semi-automated matching between narrative findings and spatial candidates. When radiology reports are available, integrating report-derived descriptors via rule-based NLP may accelerate retrospective dataset characterization while preserving a clear audit trail for human verification.

Limitations include reliance on underlying open-source model implementations and operating thresholds determined on specific validation cohorts. Tri-Reader is not intended to replace expert reading; instead, it prioritizes practical annotation efficiency and reproducibility. The pipeline was evaluated on retrospective datasets; prospective workflow studies are needed to assess real-world time savings and annotation quality. Future work includes site-specific calibration, integration of additional open-source models as they become available, expansion to longitudinal screening scenarios and adding vision-language models.

**Data and Code Availability**

Tri-Reader will be released as an open-access tool with installation scripts, pretrained weights (where licenses permit), and evaluation code at: https://github.com/fitushar/TriAnnot/

Dataset resources are available at the following repositories:

**Health AI Data Resource (HAID):** https://github.com/fitushar/HAID

**LUNA16:** https://luna16.grand-challenge.org/Data/

**LUNA25:** https://luna25.grand-challenge.org/

DLCS24: https://zenodo.org/records/13799069

**Supplementary Material**

The Supplement provides extended methodological details of the Tri-Reader pipeline and additional stratified performance results across all test cohorts, including FROC analyses, missed-nodule characterization, and detectability analyses.

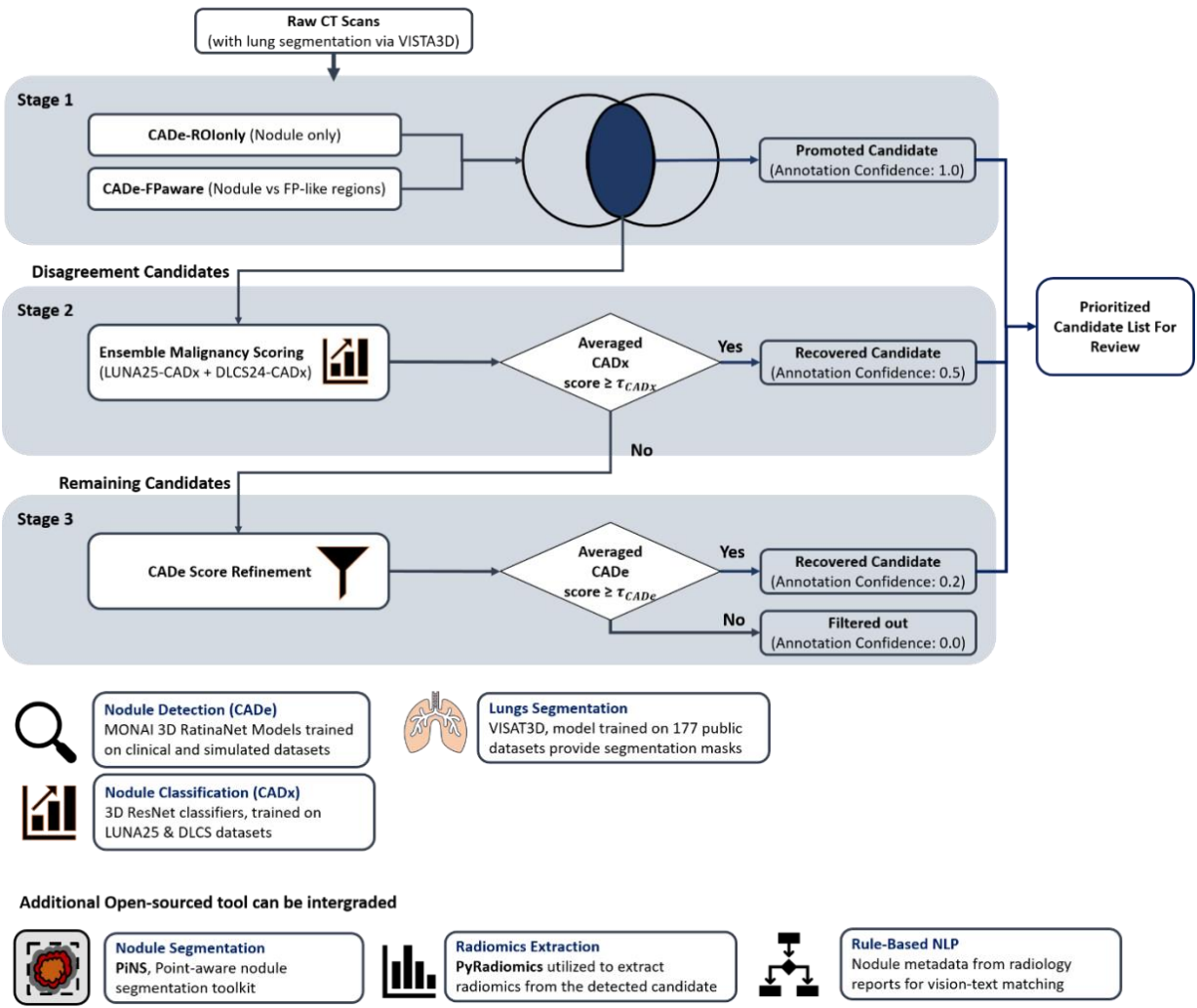

**Figure 1**. **Tri-Reader tri-stage workflow for first-pass lung nodule annotation in screening CT.** After lung segmentation (VISTA3D) removes extra-pulmonary candidates, Stage 1 performs *consensus CADe* using two complementary detection models (*CADe*-ROIonly and the false-positive–aware *CADe*-FPaware trained via strategic hard-negative mining); candidates detected by both are promoted to confidence = 1.0. **Stage 2** applies ensemble malignancy scoring to disagreement candidates using two CADx classifiers trained on complementary distributions (DLCS24-CADxr50SWS and LUNA25-CADxr50); candidates with averaged *CADx* ≥ 0.10 are promoted (confidence = 0.5). **Stage 3** retains remaining candidates meeting *CADe* ≥ 0.20 (confidence = 0.2), yielding a single merged candidate list with tiered confidence intended to reduce annotator workload while preserving sensitivity; an optional rule-based NLP module can link report-derived descriptors to candidates when radiology reports are available.

**Table 1: Demographics, Inclusion Criteria and Petitions of the datasets.**

| Dataset | Partitions | Centers, Date range | No. of Patients CT, Nodules (n) | Inclusion & Exclusion Criteria | No. of Female Patients (%), Age ± std | Annotation |
|---|---|---|---|---|---|---|
| DLCS24 Benchmark [3, 7] | Development & Test | 1 Center in USA; January 2015 to June 2021 | 1605 Patient; 1605 CTs; 2478 Nodules | Entire public benchmark data is included; No exclusion | 802 (49.72); 66 ± 6. | Nodules: AI & Radiologist Annotated Cancer Diagnosis: Histopathology & Clinical follow-up |
| LUNA16 [14] | Development | 7 academic institutions in USA; Between 2001 to 2004 | N/A Patient; 888 CTs; 1186 Nodules | Included only pre-defined 10-folds by MONAI Detection Module: 601 CTs; 1186 Nodules | N/A | Radiologist Annotated Nodules; Cancer Diagnosis: Radiologist Suspicions Score |
| LUNA25 [2] | Development | 33 Center in USA; August 2022 to December 2015 | 2120 Patient; 4069 CTs; 6163 Nodules | Entire public data is included; No exclusion | 909 (43%); 63 ± 5 | Nodules: AI & Radiologist Annotated Cancer Diagnosis: Histopathology & Clinical follow-up |
| VLST [15] | Development | 1 Center in USA; December 2023 | 174 Patients; 1044 CTs; 3072 Nodules | No exclusion is performed | 79 (45%); 59 ± 15; | |
| **Test Dataset** | | | | | | |
| DLCS Dataset-Private [3, 7] | Test | 1 Center in USA; January 2015 to June 2021 | 419 Patients; 419 CTs; 651 Nodules | No exclusion is performed | 211 (50%); 66 ± 6 | Nodules: AI & Radiologist Annotated Cancer Diagnosis: Histopathology & Clinical follow-up |
| LNDbv4 [16] | External Test | 2 Centers in Portugal; Between 2016 and 2018 | 212 Patients; 212 CTs; 768 Nodules | Included text–image concordance on and excluded rest; 155 Patients; 155 CTs; 377 Nodules | N/A | Radiologist Annotated Nodules; Cancer Diagnosis: Radiologist Suspicions Score |
| IMD-CT [4] | External Test | 5 Center in China; March 2021 to December 2021 | 174 Patients; 1044 CTs; 3072 Nodules | Entire public data is included; No exclusion | 1216 (60%); 55 ± 12 | Nodules: AI & Radiologist Annotated Cancer Diagnosis: Histopathology |

**Note:** Duke Lung Cancer Screening= DLCS24, Virtual Lungs Cancer Screening = VLST, Lung Nodule Database = LNDbv4, Integrated Multiomics = IMD-CT.

Table 2. Detection performance across internal and external test cohorts.

| Dataset & Model | CPM | Sen. @1 FP/scan | Detection Rate (Candidate/Scans) |
|---|---|---|---|
| **LNDbv4** | | | |
| LUNA16-De | 0.51 (0.46-0.56) | 0.52 (0.46-0.58) | 19.30 (2992/155) |
| Tri-Reader | **0.47 (0.42-0.52)** | **0.47 (0.41-0.54)** | **11.34 (1757/155)** |
| **IMD-CT** | | | |
| LUNA16-De | 0.63 (0.62-0.65) | 0.66 (0.64-0.69) | 13.9 (28258/2032) |
| Tri-Reader | **0.73 (0.72-0.75)** | **0.80 (0.79-0.83)** | **6.71 (13642 /2032)** |
| **DLCS Benchmark** | | | |
| LUNA16-De | 0.57 (0.53-0.62) | 0.61 (0.52-0.66) | 23.1 (4581 / 198) |
| Tri-Reader | **0.62 (0.58-0.67)** | **0.68 (0.59-0.73)** | **13.0 (2575 / 198)** |
| **DLCS Private** | | | |
| LUNA16-De | 0.53 (0.50-0.56) | 0.53 (0.49-0.59) | 22.3 (9,349 / 419) |
| Tri-Reader | **0.59 (0.56-0.62)** | **0.60 (0.55-0.65)** | **12.4 (5,208 / 419)** |

**Reference**

[1] P. G. Mikhael *et al.*, "Sybil: A Validated Deep Learning Model to Predict Future Lung Cancer Risk From a Single Low-Dose Chest Computed Tomography," *J Clin Oncol,* vol. 41, no. 12, pp. 2191-2200, Apr 20 2023, doi: 10.1200/JCO.22.01345.
[2] D. Peeters, B. Obreja, N. Antonissen, and C. Jacobs, "The LUNA25 Challenge: Public Training and Development set - Imaging Data," doi: 10.5281/zenodo.14223624.
[3] A. J. Wang *et al.*, "The Duke Lung Cancer Screening (DLCS) Dataset: A Reference Dataset of Annotated Low-Dose Screening Thoracic CT," *Radiol Artif Intell,* vol. 7, no. 4, p. e240248, Jul 2025, doi: 10.1148/ryai.240248.
[4] M. Zhao *et al.*, "Integrated multiomics signatures to optimize the accurate diagnosis of lung cancer," *Nat Commun,* vol. 16, no. 1, p. 84, Jan 2 2025, doi: 10.1038/s41467-024-55594-z.
[5] Y. He *et al.*, "VISTA3D: A unified segmentation foundation model for 3D medical imaging," in *Proceedings of the Computer Vision and Pattern Recognition Conference*, 2025, pp. 20863-20873.
[6] F. Isensee, P. F. Jaeger, S. A. A. Kohl, J. Petersen, and K. H. Maier-Hein, "nnU-Net: a self-configuring method for deep learning-based biomedical image segmentation," *Nature Methods,* vol. 18, no. 2, pp. 203-211, 2021, doi: 10.1038/s41592-020-01008-z.
[7] F. I. Tushar *et al.*, "AI in Lung Health: Benchmarking Detection and Diagnostic Models Across Multiple CT Scan Datasets," *arXiv preprint arXiv:2405.04605,* 2024.
[8] M. J. Cardoso *et al.*, "Monai: An open-source framework for deep learning in healthcare," *arXiv preprint arXiv:2211.02701,* 2022.
[9] M. Baumgartner, P. F. Jäger, F. Isensee, and K. H. Maier-Hein, "nnDetection: a self-configuring method for medical object detection," in *Medical Image Computing and Computer Assisted Intervention–MICCAI 2021: 24th International Conference, Strasbourg, France, September 27–October 1, 2021, Proceedings, Part V 24*, 2021: Springer, pp. 530-539.

[10] S. Pai *et al.*, "Foundation model for cancer imaging biomarkers," *Nature Machine Intelligence,* vol. 6, no. 3, pp. 354-367, 2024/03/01 2024, doi: 10.1038/s42256-024-00807-9.
[11] S. Chen, K. Ma, and Y. Zheng, "Med3d: Transfer learning for 3d medical image analysis," *arXiv preprint arXiv:1904.00625,* 2019.
[12] Z. Zhou, V. Sodha, J. Pang, M. B. Gotway, and J. Liang, "Models Genesis," *Med Image Anal,* vol. 67, p. 101840, Jan 2021, doi: 10.1016/j.media.2020.101840.
[13] V. Sundaram, M. K. Gould, and V. S. Nair, "A Comparison of the PanCan Model and Lung-RADS to Assess Cancer Probability Among People With Screening-Detected, Solid Lung Nodules," *Chest,* vol. 159, no. 3, pp. 1273-1282, 2021/03/01/ 2021, doi: https://doi.org/10.1016/j.chest.2020.10.040.
[14] A. A. A. Setio *et al.*, "Validation, comparison, and combination of algorithms for automatic detection of pulmonary nodules in computed tomography images: The LUNA16 challenge," *Med Image Anal,* vol. 42, pp. 1-13, Dec 2017, doi: 10.1016/j.media.2017.06.015.
[15] F. I. Tushar *et al.*, "Virtual Lung Screening Trial (VLST): An In Silico Study Inspired by the National Lung Screening Trial for Lung Cancer Detection," *Medical Image Analysis,* p. 103576, 2025.
[16] C. A. Ferreira *et al.*, "LNDb v4: pulmonary nodule annotation from medical reports," *Sci Data,* vol. 11, no. 1, p. 512, May 17 2024, doi: 10.1038/s41597-024-03345-6.

# Supplements

# Tri-Reader: An Open-Access, Multi-Stage AI Pipeline for First-Pass Lung Nodule Annotation in Screening CT

**Fakrul Islam Tushar, PhD, Joseph Y. Lo PhD**

Center for Virtual Imaging Trials, Carl E. Ravin Advanced Imaging Laboratories, Department of Radiology, Duke University School of Medicine, Durham, NC

Dept. of Electrical & Computer Engineering, Pratt School of Engineering, Duke University, Durham

This Supplement provides supplementary methods, additional experiments, and extended quantitative results referenced in the main manuscript.

## Table of Contents

# Appendix S1: Extended Pipeline Implementation Details

## S1.1 Detection Model (CADe) Specifications

The detection stage in Tri-Reader is designed to identify all potential nodule candidates through, **CADe-ROIonly** and **CADe-FPaware** detection models. **CADe-ROIonly** detection model was trained to localize pulmonary nodules using four curated CT datasets: DLCS24 Benchmark dataset [1], LUNA16 [2], LUNA25 [3], and the VLST simulated CT dataset [4]. The VLST simulated dataset includes 174 digital human twins with 512 simulated lung nodules, yielding 1,044 CT scans and 3,072 nodule images across two scanners. The cohort comprised 95 males (54.6%), with mean age 59 ± 15 years and BMI 26 ± 6. Demographics were 73.6% White, 20.7% Black, and 5.8% Other/Unknown, with 98.3% non-Hispanic. Simulation methods and dataset construction are described in prior work [4, 5]. Detailed descriptions of the DLCS24 and LUNA25 datasets and data splits are provided in **Tables S1.1** and **S1.2**.

All volumes were resampled to uniform voxel spacing and intensity-normalized, and annotations were harmonized across datasets. A 3D RetinaNet architecture implemented in MONAI was used to predict bounding boxes and confidence scores for candidate nodules. **CADe-FPaware** detection model were trained utilizing strategic hard negatives sampled from the above datasets proposed in a prior study [1]. After the baseline detector was trained, inference was performed on the same datasets to generate full-volume prediction maps. From these results, strategic hard negatives were identified as regions consistently predicted as nodules but confirmed as non-nodular through reference labels [1]. These hard negatives were added to the original positive set to create an augmented dataset containing both true nodules and visually similar non-nodular structures. **CADe-FPaware** detection model, identical in architecture, was trained on this combined dataset. Incorporating hard negatives enabled the network to learn more discriminative features and reduced the overall false-positive rate.

The model training and evaluation pipeline was adapted from an open-source implementation (GitHub repository: *AI-in-Lung-Health-Benchmarking Detection and Diagnostic Models Across Multiple CT Scan Datasets*). available at: https://github.com/fitushar/AI-in-Lung-Health-Benchmarking-Detection-and-Diagnostic-Models-Across-Multiple-CT-Scan-Datasets

**Table S1.1. DLCS24 benchmark cohort characteristics and data splits.** Counts of patients, CT scans, and nodule annotations with age (mean ± SD), sex distribution, mean nodule size (mm), and diagnosis labels summarized for the full cohort and each split.

| Category | Value | Total | Train | Validation | Test |
|---|---|---|---|---|---|
| **Patient** | Patients | 1605 (100.0%) | 1054 (65.7%) | 354 (22.1%) | 197 (12.3%) |
| **CT** | CT scans | 1605 (100.0%) | 1054 (65.7%) | 354 (22.1%) | 197 (12.3%) |
| **Age (years)** | Mean ± SD | 66.8 ± 6.1 | 66.9 ± 6.2 | 66.4 ± 6.0 | 67.0 ± 5.9 |
| **Gender** | Female | 1251 (50.5%) | 776 (48.1%) | 304 (53.1%) | 171 (58.4%) |
| | Male | 1227 (49.5%) | 837 (51.9%) | 268 (46.9%) | 122 (41.6%) |
| **Nodule** | Annotation Count | 2478 (100.0%) | 1613 (65.1%) | 572 (23.1%) | 293 (11.8%) |
| | Mean ± SD (mm) | 12.3 ± 8.9 | 12.6 ± 9.6 | 11.8 ± 7.6 | 11.7 ± 7.0 |
| **Diagnosis** | Benign | 2214 (89.3%) | 1447 (89.7%) | 507 (88.6%) | 260 (88.7%) |
| | Cancer | 264 (10.7%) | 166 (10.3%) | 65 (11.4%) | 33 (11.3%) |

**Table S1.2. LUNA25 cohort characteristics and train/validation/test splits.** Counts of patients, CT scans, and nodule annotations with age (mean ± SD), sex distribution, mean nodule size (mm), and diagnosis labels summarized for the full cohort and each split.

| Category | Value | Total | Train | Validation | Test |
|---|---|---|---|---|---|
| **Patient** | Patients | 1993 (100.0%) | 1594 (80.0%) | 199 (10.0%) | 200 (10.0%) |
| **CT** | CT scans | 3792 (100.0%) | 3039 (80.1%) | 389 (10.3%) | 364 (9.6%) |
| **Age (years)** | Mean ± SD | 63.3 ± 5.3 | 63.4 ± 5.3 | 63.0 ± 4.9 | 63.4 ± 5.6 |
| **Gender** | Female | 2414 (40.9%) | 1941 (40.9%) | 216 (37.8%) | 257 (44.1%) |
|  | Male | 3485 (59.1%) | 2804 (59.1%) | 355 (62.2%) | 326 (55.9%) |
| **Nodule** | Annotation Count | 5899 (100.0%) | 4745 (80.4%) | 571 (9.7%) | 583 (9.9%) |
|  | Mean ± SD (mm) | 12.7 ± 6.8 | 12.7 ± 6.9 | 13.1 ± 6.4 | 12.5 ± 6.6 |
| **Diagnosis** | Benign | 5450 (92.4%) | 4393 (92.6%) | 523 (91.6%) | 534 (91.6%) |
|  | Cancer | 449 (7.6%) | 352 (7.4%) | 48 (8.4%) | 49 (8.4%) |

### S1.2 Lung Segmentation Integration

In our pipeline, lung segmentation is performed automatically using VISTA3D, a 3D foundation segmentation model that supports 127 anatomical classes and is readily accessible through MONAI [6]. VISTA3D employs a 3D SegResNet backbone with sliding-window inference (128×128×128 voxel patches) to preserve full volumetric context during segmentation. Reported results on the TotalSegmentatorV2 test split demonstrate strong lobe-level performance, with Dice scores of 0.95 (left upper lobe), 0.94 (left lower lobe), 0.88 (right upper lobe), 0.92 (right middle lobe), and 0.943 (right lower lobe), indicating accuracy comparable to specialized expert models, such as nnUNet and Auto3dSeg [6-8]. For Tri-Reader, we generate lung (lobe IDs 28: left upper, 29: left lower, 30: right upper, 31: right middle, 32: right lower) masks and use them to reject candidates outside the lungs, ensuring downstream *CADe*/*CADx* analysis is restricted to anatomically plausible regions; our primary use is automated masking for candidate filtering. We selected VISTA3D due to its strong reported performance and its well-documented, open-source implementation within MONAI [6]. Importantly, Tri-Reader is modular, and the segmentation component can be replaced or updated as needed.

### S1.3 Malignancy Classification (CADx) Model Specifications

#### S1.3.1 LUNA25-CADxr50 Architecture

A 3D ResNet50 classifier was trained on the LUNA25 development dataset cohort to predict nodule malignancy. We adopted the 3D ResNet50 CAD*x* framework from a prior study [1]. The classifier was implemented with one input channel and two output classes. Training was performed for 200 epochs with a batch size of 24 for both training and validation, with validation run every 5 epochs. We did not apply sampling-based rebalancing and optimized the network with an initial learning rate of $1 \times 10^{-2}$. Training was conducted from scratch. The LUNA25 cohort was split into training/validation/test, comprising 1594/199/200 patients, 3039/389/364 CT scans, and 4745/571/583 nodule annotations, with 352/48/49 cancer-positive nodules in the respective splits. The final checkpoint was selected based on the best validation-set performance. We addressed class imbalance in LUNA25 by using class-weighted cross-entropy loss. We computed inverse-frequency class weights from the distribution of ground-truth labels in the training split. This increases the penalty for misclassifying the minority cancer class relative to the majority benign class. During training, the objective for each sample with a label was:

$$\mathcal{L}(x, y) = -w_y \log p_y(x)$$

where $p_y(x)$ is the SoftMax probability assigned by the model to the true class.

The model training and evaluation pipeline was adapted from an open-source implementation (GitHub repository: *AI-in-Lung-Health-Benchmarking Detection and Diagnostic Models Across Multiple CT Scan Datasets*). available at: https://github.com/fitushar/AI-in-Lung-Health-Benchmarking-Detection-and-Diagnostic-Models-Across-Multiple-CT-Scan-Datasets

### S1.3.2 DLCS24-CADxr50SWS (Strategic Warm-Start)

We adopted the ResNet50-SWS from the prior DLCS24 benchmarking study [1]. Strategic Warm-Start (SWS) uses a three-stage procedure: (i) candidate curation, where 64×64×64 voxel patches are assembled from the DLCS24 benchmark training and validation cohort using annotated nodules as positives and detector-derived false positives as negatives, with negatives stratified by detector confidence (low/medium/high confidence bins) to enrich hard negatives; (ii) pretraining, where a randomly initialized 3D ResNet50 is trained to classify nodule vs non-nodule on the curated candidate set; and (iii) task-specific fine-tuning, where the pretrained weights are transferred to initialize a downstream malignancy classifier (benign vs malignant) and fine-tuned end-to-end, with the final checkpoint selected by validation performance [1]. This warm-start is intended to learn nodule-relevant representations directly from detection-driven candidates, improving false-positive suppression and stabilizing malignancy learning when large-scale external pretraining is unavailable.

The model evaluation pipeline and weights were adapted from an open-source implementation (GitHub repository: *AI-in-Lung-Health-Benchmarking Detection and Diagnostic Models Across Multiple CT Scan Datasets*). available at: https://github.com/fitushar/AI-in-Lung-Health-Benchmarking-Detection-and-Diagnostic-Models-Across-Multiple-CT-Scan-Datasets

### S1.3.3 Ensemble averaging (Tri-Reader Stage 2)

For both *CADx* models, candidate-centered patches were generated using the nodule centroid (x, y, z) from the detection stage. A 64×64×64-voxel cube was extracted around each centroid, resampled to 0.7×0.7×1.25 mm spacing, and intensity standardized using HU windowing (−1000 to 500 HU) followed by normalization. Patches extending beyond the scan boundary were handled using zero-padding. All patch generation, preprocessing, and scoring steps are fully automated within the Tri-Reader pipeline, requiring no manual intervention once candidate coordinates are available from Stage-1 of Tri-Reader.

The LUNA25-trained CADx (LUNA25-CADxr50) model and the DLCS24 (DLCS24CADxr50-SWS) *CADx* model each produced an independent malignancy probability (0,1). The final ensemble score was computed as the unweighted mean:

$$Avg.CADx = \frac{CADx_{LUNA25}\,p_y(x) + CADx_{DLCS24-SWS}\,p_y(x)}{2}$$

where $p_y(x)$ is the SoftMax probability assigned by the respective models. This averaging combines complementary behavior across models.

### S1.4 Operating Threshold Selection Methodology

### S1.4.1 *CADx* Operating Threshold Selection

We selected the average *CADx* operating threshold empirically from the validation analysis shown in **Figure S1.1** by sweeping $\tau_{CADx}$ and examining the resulting sensitivity-workload trade-off. We fixed $\tau_{CADx} = 0.10$ because it provided a high-sensitivity screening operating point with a recall 0.95 (397/417 cancers retained; 20 false negatives) while maintaining moderate enrichment (PPV 0.35) and a manageable reduction in low-risk candidates (725 false positives; FPR 0.63; 71.83% of candidates forwarded). This threshold lies at the onset of separation between the non-cancer and cancer score distributions (density/CDF), capturing most cancer-associated candidates while filtering a subset of low-scoring non-cancer candidates. The selected $\tau_{CADx}$ was then fixed and applied unchanged in the Tri-Reader pipeline.

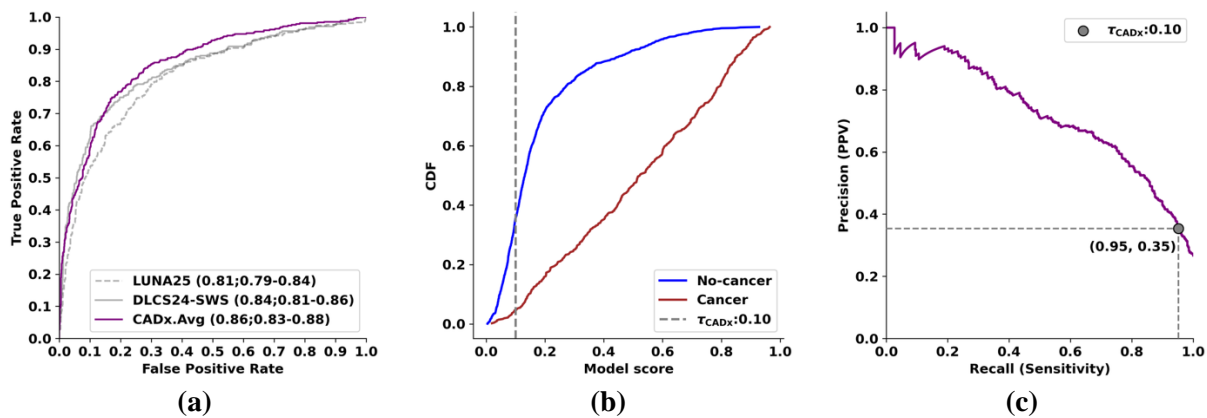

**Figure S1.1.** Ensemble CADx performance and operating-threshold selection for Tri-Reader Stage 2. **(a)** Receiver operating characteristic (ROC) curves comparing the LUNA25-trained CADx model, the DLCS24-SWS CADx model, and their unweighted ensemble (**CADx.Avg**). Area under the ROC curve (AUC) values are shown with 95% confidence intervals. **(b)** Empirical cumulative distribution functions (CDFs) of ensemble CADx scores for no-cancer and cancer candidates, illustrating separation between groups at low score ranges. **(c)** Precision–recall (PR) curve for the ensemble CADx score. The dashed vertical line and marker denote the selected operating point, $\tau_{CADx} = 0.10$, corresponding to **recall 0.95** and **precision 0.35** (**Table S1.3**). Candidates and centroids are inherited from **Tri-Reader Stage 1**, and all patch extraction, preprocessing, and CADx scoring are performed automatically within Tri-Reader.

**Table S1.3. Sensitivity-workload trade-off across *CADx* thresholds used for Tri-Reader Stage 2 filtering.** Threshold sweep results for the ensemble *CADx* score. "Missed" denotes the number of cancer cases missed at each threshold (false negatives, FN). "Flagged (%)" denotes the percentage of candidates forwarded to subsequent Tri-Reader processing (score ≥ threshold). Recall and precision are reported for the cancer class. FPR denotes the false-positive rate among no-cancer candidates. FN, FP, and TP denote false negatives, false positives, and true positives, respectively. The prespecified operating point $\tau_{CADx} = 0.10$ yields a recall of 0.95 (TP=397, FN=20) with a precision of 0.35, while forwarding 71.83% of candidates (FP=725).

| Missed | Threshold | Recall | Precision | FPR | Flagged (%) | FN | FP | TP |
|---|---|---|---|---|---|---|---|---|
| 0 | 0.02 | 1.00 | 0.27 | 0.98 | 98.21 | 0 | 1117 | 417 |
| 1 | 0.03 | 1.00 | 0.27 | 0.97 | 97.44 | 1 | 1106 | 416 |
| 2 | 0.03 | 1.00 | 0.27 | 0.96 | 96.93 | 2 | 1099 | 415 |
| 3 | 0.04 | 0.99 | 0.28 | 0.94 | 95.58 | 3 | 1079 | 414 |
| 5 | 0.04 | 0.99 | 0.28 | 0.92 | 93.47 | 5 | 1048 | 412 |
| 10 | 0.08 | 0.98 | 0.32 | 0.76 | 81.50 | 10 | 866 | 407 |
| 20 | **0.10** | 0.95 | 0.35 | 0.63 | 71.83 | 20 | 725 | 397 |

## S1.4.2 *CADe* Operating Threshold Selection

We selected the *CADe* minimum score threshold $\tau_{CADe} = 0.20$ based on a validation threshold sweep ($\tau_{CADe} = 0.05\text{-}0.50$), evaluating the sensitivity-workload trade-off using FROC analysis across cohorts. Across datasets, average FROC sensitivity was relatively stable across thresholds, whereas the number of candidates forwarded downstream decreased monotonically with increasing $\tau_{CADe}\ threshold$. We selected $\tau_{CADe} = 0.20$ a conservative screening operating point that reduced low-confidence candidate volume while maintaining high sensitivity, avoiding the disproportionate increase in missed nodules observed at higher thresholds. This threshold was then fixed and used for all Tri-Reader Stage-3 candidate generation. **Table S1.4** reports the *CADe* score threshold sweep on the **validation split** ($\tau_{CADe} = 0.05\text{-}0.50$), including average FROC sensitivity (AvgSens), the number of candidates forwarded downstream, and the number of missed reference nodules. Increasing $\tau_{CADe}$ reduces candidate volume monotonically while increasing missed nodules, whereas AvgSens remains relatively stable across thresholds. We selected $\tau_{CADe} = 0.20(†)$ as a conservative operating point balancing sensitivity with downstream workload, and this threshold was fixed for all Tri-Reader Stage-1 experiment.

**Table S1.4. CADe threshold on validation split.** CPM = average sensitivity; Candidates = detections forwarded downstream; Missed = reference nodules not detected at the specified threshold. † marks the selected operating point $\tau_{CADe} = 0.20$.

| Dataset | $\tau_{CADe}$ | CPM | Candidates (n) | Missed (n) |
|---|---|---|---|---|
| DLCS24 | 0.05 | 0.643 | 4168 | 29 |
| DLCS24 | 0.10 | 0.643 | 3538 | 30 |
| DLCS24 | 0.15 | 0.650 | 3212 | 34 |
| **DLCS24** | **0.20†** | **0.650** | **3050** | **35** |
| DLCS24 | 0.25 | 0.650 | 2883 | 40 |
| DLCS24 | 0.30 | 0.650 | 2718 | 42 |
| DLCS24 | 0.35 | 0.650 | 2625 | 45 |
| DLCS24 | 0.40 | 0.650 | 2555 | 46 |
| DLCS24 | 0.45 | 0.650 | 2433 | 48 |
| DLCS24 | 0.50 | 0.650 | 2335 | 48 |
| LUNA16 | 0.05 | 0.780 | 653 | 2 |
| LUNA16 | 0.10 | 0.782 | 557 | 3 |
| LUNA16 | 0.15 | 0.782 | 509 | 3 |
| **LUNA16** | **0.20†** | **0.782** | **489** | **3** |
| LUNA16 | 0.25 | 0.782 | 473 | 3 |
| LUNA16 | 0.30 | 0.782 | 453 | 3 |
| LUNA16 | 0.35 | 0.782 | 449 | 3 |
| LUNA16 | 0.40 | 0.782 | 439 | 3 |
| LUNA16 | 0.45 | 0.782 | 425 | 3 |
| LUNA16 | 0.45 | 0.782 | 425 | 3 |
| Simulated | 0.05 | 0.966 | 1428 | 13 |
| Simulated | 0.10 | 0.970 | 1197 | 15 |
| Simulated | 0.15 | 0.967 | 1119 | 15 |
| **Simulated** | **0.20†** | **0.971** | **1035** | **15** |
| Simulated | 0.25 | 0.971 | 1008 | 15 |
| Simulated | 0.30 | 0.971 | 990 | 15 |
| Simulated | 0.35 | 0.971 | 975 | 16 |
| Simulated | 0.40 | 0.971 | 964 | 16 |
| Simulated | 0.45 | 0.971 | 936 | 19 |
| Simulated | 0.50 | 0.971 | 912 | 19 |
| Combined | 0.05 | 0.796 | 6249 | 46 |
| Combined | 0.10 | 0.796 | 5292 | 48 |
| Combined | 0.15 | 0.802 | 4840 | 52 |
| **Combined** | **0.20†** | **0.802** | **4521** | **53** |
| Combined | 0.25 | 0.802 | 4365 | 58 |
| Combined | 0.30 | 0.802 | 4161 | 61 |
| Combined | 0.35 | 0.803 | 3844 | 66 |
| Combined | 0.40 | 0.803 | 3655 | 71 |
| Combined | 0.45 | 0.806 | 3428 | 78 |
| Combined | 0.50 | 0.811 | 3356 | 84 |

**Appendix S3: Statistical Analysis Methods**

**S3.1 FROC Analysis**

**Lesion-level matching criteria.** Detection performance was evaluated using lesion-level matching, where a predicted candidate was counted as a true positive if the candidate centroid was within a prespecified Euclidean distance of a reference nodule centroid. One-to-one matching was enforced such that each reference nodule could be assigned to at most one predicted candidate. **Distance tolerance.** Following the LUNA16 evaluation protocol, the matching tolerance was diameter-dependent to account for size-related annotation variability: for nodules <10 mm, the allowable distance was set to one-half of the nodule diameter, whereas for nodules ≥10 mm, the tolerance was capped at 5 mm. This diameter-dependent tolerance accounts for annotation uncertainty, which scales with nodule size.

**CPM Calculation.** The Competition Performance Metric (CPM) [2] is defined as the average sensitivity across seven predefined false positive (FP) rates per scan:

$$\text{CPM} = \frac{1}{7}\sum_{i=1}^{7} \text{Sensitivity}(\text{FP}_i)$$

where FP rates are: 1/8, 1/4, 1/2, 1, 2, 4, 8 FPs per scan. For each FP rate threshold $t$:

a. Rank all candidates by detection score (descending)
b. Determine the score threshold $\tau$ such that total FPs per scan = $t$
c. Count TPs at threshold $\tau$
d. Sensitivity ($t$) = TPs / total ground-truth nodules

**S3.2 Radiologist Consensus Analysis Methods (LNDbv4)**

**Radiologist Vote Pattern Categorization.** Lung Nodule Database (LNDbv4)[9], provides multi-reader annotations with vote counts indicating how many radiologists identified each nodule. We categorized patterns as:

- **1R (1/1):** Single-reader annotation (1 radiologist reviewed, 1 marked nodule)
- **2R (2/2) Consensus:** Two radiologists reviewed, both marked nodule
- **2R (1/2) Disagree:** Two radiologists reviewed, only one marked nodule

- **3R (3/3) Consensus:** Three radiologists reviewed, all three marked nodule
- **3R (2/3) 1 dissent:** Three radiologists reviewed, two marked nodule (one dissented)

These patterns serve as proxy measures of nodule conspicuity and clinical significance.

**Cohen's d effect size calculations [10].** For semantic characteristics comparing detected vs missed nodules, we computed standardized effect sizes:

$$d = \frac{\bar{X}_{\text{detected}} - \bar{X}_{\text{missed}}}{s_{\text{pooled}}}$$

where pooled standard deviation:

$$s_{\text{pooled}} = \sqrt{\frac{(n_1 - 1)s_1^2 + (n_2 - 1)s_2^2}{n_1 + n_2 - 2}}$$

Effect size interpretation:

- Small: $|d| = 0.2–0.5$
- Medium: $|d| = 0.5–0.8$
- Large: $|d| > 0.8$

**Mann-Whitney U Test Procedures [11].** For ordinal semantic ratings (Subtlety, Malignancy, Texture, etc.), we used non-parametric Mann-Whitney U tests to compare detected vs missed distributions:

- Null hypothesis: Distributions are identical
- Alternative hypothesis: Two-sided (detected ≠ missed)
- Significance threshold: $\alpha = 0.05$

**Appendix S4: Test Dataset Detailed Results**

**S4.1 DLCS24 Test Dataset Detailed Results**

The Duke Lung Cancer Screening (DLCS24) Test set comprised 198 CT scans from 198 patients with 294 annotated nodules [1, 12]. Mean patient age was 66 years (range 54-79), with 42.9% male and 57.1% female. Most patients were White (70.7%) or Black/African American (25.8%). At the nodule level, 11.2% (33/294) were pathologically confirmed malignancies. **Table S4.1**detailed the demographics and nodule characteristics in the DLCS24 benchmark test cohort.

**S4.1.1 Overall and Size-Stratified Detection Performance**

**Table S4.2** presents FROC results stratified by nodule size. Across all nodules, Tri-Reader achieved an average sensitivity/CPM 0.62 (95% CI: 0.58-0.67) with 13.0 candidates per scan, representing a notable reduction compared to LUNA16-De (23.1 candidates/scan) while maintaining statistically equivalent sensitivity (overlapping 95% CIs). Size-stratified analysis revealed expected performance scaling with nodule diameter. For 6 mm nodules (n=167, 56.8% of cohort), both models showed limited sensitivity at low FP rates, with Tri-Reader achieving a sensitivity of 0.54 at 1 FP/scan and 0.87 at 4 FP/scan. For 6-10 mm nodules (n=90, 30.6%), Tri-Reader maintained competitive performance (sensitivity 0.64 at 1 FP/scan) while substantially reducing candidate burden (15 vs 27 candidates/scan for LUNA16-De). For nodules ≥10 mm (n=37, 12.6%), Tri-Reader achieved high sensitivity (0.89 at 4 FP/scan, 0.97 at 8 FP/scan) with only 14.4 candidates/scan.

**S4.1.2 Lung-RADS Category Performance**

**Table S4.3** stratifies detection performance by Lung-RADS v2022 category. For Lung-RADS 4B nodules (n=35), the Tri-Reader achieved a 100% detection rate (35/35), exceeding the LUNA16-De (34/35). In this category, the Tri-Reader's median detection probability was 0.98, with a minimum probability of 0.26, suggesting significantly higher detection confidence compared to the LUNA16-De baseline (min probability 0.04). For Lung-RADS 3 nodules (n=50), the Tri-Reader identified 47/50 (94%) of lesions, with a mean probability of 0.89. In the Lung-RADS 4X category (n=8), representing nodules with the highest suspicion of malignancy, the Tri-Reader identified 7 out of 8 nodules with a high mean detection probability of 0.98. The distribution for the Tri-Reader in this category was notably tight (SD = 0.02), reinforcing the model's stability when evaluating radiologically concerning findings.

**S4.1.3 Missed Nodule Characteristics**

**Table S4.4** summarizes the characteristics of nodules missed by each detection model. Tri-Reader missed 12 nodules total (4.1% of cohort), of which 7 (58.3%) were uniquely missed (not missed by LUNA16-De). These unique misses had small median diameter 5.90 mm (range 2.92-17.19 mm), with 6/7 (85.7%) being 6 mm nodules (Lung-RADS 2). In contrast, LUNA16-De unique misses (n=6) were substantially larger (mean diameter 12.32 mm), including a 21.97 mm mass that was successfully detected by Tri-Reader.

**Table S4.1. Demographics and Nodule Characteristics in the DLCS24 Benchmark Test Cohort.**

| Category | Sub-category | Test Dataset | |
|---|---|---|---|
| **Age** | Mean (Min- Max) | 66 (55-82) | |
| **Gender** | Male | 85 (42.93%) | |
| | Female | 113 (57.07%) | |
| **Race** | White (ref) | 140 (70.71%) | |
| | Black/AA | 51 (25.76%) | |
| | Other/Unknown | 7 (3.54%) | |
| **Ethnicity** | Not Hispanic | 192 (96.97%) | |
| | Unavailable | 5 (2.53%) | |
| | Hispanic | 1 (0.51%) | |
| **Smoking status** | Currently smoking | 99 (52.38%) | |
| | Formerly smoking | 90 (47.62%) | |
| **Tobacco used years** | 30+ years | 138 (79.31%) | |
| | 21-30 years | 31 (17.82%) | |
| | 11-20 years | 3 (1.72%) | |
| | 0-5 years | 2 (1.15%) | |
| | | **Patient** | **Nodule** |
| **Malignancy** | Benign | 180 (90.91%) | 261 (88.78%) |
| | Cancer | 18 (9.09%) | 33 (11.22%) |
| **Lung-RADS** | 2 | 113 (57.07%) | 152 (51.70%) |
| | 3 | 34 (17.17%) | 50 (17.01%) |
| | 4A | 28 (14.14%) | 48 (16.33%) |
| | 4B | 17 (8.59%) | 35 (11.90%) |
| | 4 | 5 (2.53%) | 8 (2.72%) |
| | 1 | 1 (0.51%) | 1 (0.34%) |
| **Lung Cancer Type** | Adenocarcinoma | 7 (41.18%) | 15 (46.88%) |
| | Squamous cell carcinoma | 6 (35.29%) | 13 (40.62%) |
| | NSCC (other/NOS) | 1 (5.88%) | 1 (3.12%) |
| | Small cell carcinoma | 3 (17.65%) | 3 (9.38%) |
| **LCS Detected Cancer** | 1 / 1.0 | 13 (76.47%) | 25 (78.12%) |
| | 0 / Unknown | 4 (23.53%) | 7 (21.88%) |
| **Nodule Size** | < 6 mm | | 167 (56.8%) |
| | 6–10 mm | | 90 (30.6%) |
| | 10–20 mm | | 28 (9.86 %) |
| | > 20 mm | | 8 (2.72%) |

**Table S4.2. Free-response Receiver Operating Characteristic (FROC) Performance and Detection Efficiency.** Performance metrics for the proposed Tri-Reader model compared to LUNA16-De baselines across standardized nodule size bins. Avg. Sensitivity is reported with 95% CI; @ X FP indicates sensitivity at 0.5, 1, and 2 FP/scan. Detected/Lesions reports unique nodules detected; Candidates/Scans denotes total candidates per scans.

| Size Category | Model | Avg. Sensitivity | Detected/Lesions (%) | Candidates / Scans (Avg.) |
|---|---|---|---|---|
| **Overall** | LUNA16-De | 0.57 (0.53-0.62) | 283/294 (96.3%) | 4581 / 198 (23.1) |
|  | Tri-Reader | 0.62 (0.58-0.67) | 282/294 (95.9%) | 2575 / 198 (13.0) |
| **< 6 mm** | LUNA16-De | 0.50 (0.44-0.56) | 163/167 (97.6%) | 3057 / 137 (22.3) |
|  | Tri-Reader | 0.53 (0.47-0.58) | 161/167 (96.4%) | 1626 / 137 (11.9) |
| **6–10 mm** | LUNA16-De | 0.59 (0.51-0.67) | 86/90 (95.6%) | 1937 / 71 (27.3) |
|  | Tri-Reader | 0.63 (0.54-0.71) | 85/90 (94.4%) | 1125 / 71 (15.9) |
| **> 10 mm** | LUNA16-De | 0.46 (0.36-0.59) | 34/37 (91.9%) | 794 / 35 (22.7) |
|  | Tri-Reader | 0.60 (0.49-0.72) | 36/37 (97.3%) | 503 / 35 (14.4) |

**Table S4.3. Performance Metrics and Detection Probability Distribution across Lung-RADS Categories.** GT Count (n) denotes total ground-truth nodules; Detected (n) denotes nodules localized by the model. Mean, Median, and Min-Max summarize the detection probability distribution for each category.

| Lung-RADS Category | Model | GT Count (n) | Detected (n) | Probability | | |
|---|---|---|---|---|---|---|
|  |  |  |  | Mean | Median | Min–Max |
| **1** | Tri-Reader | 1 | 1 | 0.995 | 0.995 | 0.99-0.99 |
| **2** | LUNA16-De | 152 | 150 | 0.906 | 0.988 | 0.06-1.00 |
|  | Tri-Reader | 152 | 150 | 0.879 | 0.979 | 0.03-1.00 |
| **3** | LUNA16-De | 50 | 46 | 0.891 | 0.986 | 0.16-0.99 |
|  | Tri-Reader | 50 | 47 | 0.899 | 0.983 | 0.05-0.99 |
| **4A** | LUNA16-De | 48 | 44 | 0.849 | 0.985 | 0.05-1.00 |
|  | Tri-Reader | 48 | 42 | 0.929 | 0.993 | 0.21-1.00 |
| **4B** | LUNA16-De | 35 | 34 | 0.873 | 0.978 | 0.04-0.99 |
|  | Tri-Reader | 35 | 35 | 0.935 | 0.984 | 0.26-0.99 |
| **4X** | LUNA16-De | 8 | 8 | 0.959 | 0.991 | 0.79-0.99 |
|  | Tri-Reader | 8 | 7 | 0.976 | 0.974 | 0.94-0.99 |

**Table S4.4. Missed Nodule Characteristics and Overlap Analysis on DLCS24 test dataset.**

| Model | Total Missed (n) | Uniquely Missed (n) | Mean Diameter (mm) | Median Diameter (mm) | Diameter Range (mm) |
|---|---|---|---|---|---|
| **LUNA16-De** | 11 | 6 | 9.59 ± 6.05 | 6.80 | 3.78–21.97 |
| **Tri-Reader** | 12 | 7 | 6.39 ± 3.72 | 5.90 | 2.92–17.19 |

## S4.2 DLCS Private Test Dataset

The DLCS Private Test set comprised 419 CT scans from a held-out internal cohort with 651 annotated nodules [1, 12]. Mean patient age was 66 years (range 54-79), with 49.6% male and 50.4% female. Racial distribution was similar to the DLCS24 test set (72.3% White, 24.3% Black/African American). At the nodule level, 10.6% (69/651) were pathologically confirmed malignancies. **Table S4.5** detailed the demographics and nodule characteristics in the DLCS24 benchmark secret test cohort.

### S4.2.1 Overall and Size-Stratified Detection Performance

**Table S4.6** shown FROC results. Tri-Reader achieved an average sensitivity 0.59 with 12.4 candidates per scan, again demonstrating ~2× reduction in candidate burden compared to LUNA16-De (22.3 candidates/scan) while maintaining comparable sensitivity. Size-stratified analysis showed consistent trends with the DLCS24 test dataset (**Supplement Section S4.1**). For 6 mm nodules (n=413, 63.4%), Tri-Reader achieved a sensitivity 0.53 at 1 FP/scan and 0.86 at 4 FP/scan. For 6–10 mm nodules (n=150, 23.0%), sensitivity was 0.65 at 1 FP/scan with only 15.3 candidates/scan. For nodules ≥10 mm (n=88, 13.5%), sensitivity reached 0.84 at 4 FP/scan and 0.92 at 8 FP/scan.

### S4.2.2 Lung-RADS Category Performance

**Table S4.7** shown detection performance by Lung-RADS category for the Secret Test set. Consistent with DLCS24 benchmark test dataset results [1, 12], Tri-Reader demonstrated strong performance for high-risk categories: Lung-RADS 4X (n=24): 91.7% detection (22/24) with mean probability $0.98 \pm 0.05$, Lung-RADS 4B (n=75): 93.3% detection (70/75) with mean probability $0.92 \pm 0.16$, and Lung-RADS 4A (n=89): 83.1% detection (74/89) with mean probability $0.89 \pm 0.23$. For Lung-RADS 3 nodules (n=129), Tri-Reader identified 92.2% (119/129) with high median probability 0.976. The 10 missed Lung-RADS 3 nodules had mean diameter 5.4 mm, all near the 6 mm threshold. The tight probability distributions for high-risk categories (4A, 4B, 4X) with high minimum values (e.g., 4X minimum 0.81) indicate that Tri-Reader consistently assigns high confidence to suspicious lesions requiring immediate clinical action.

### S4.2.3 Diagnostic Correlation to performances

**Table S4.8** shown detection stratified by histopathological diagnosis. For the 69 malignancies, Tri-Reader detected 63 (91.3%) with high median confidence 0.99 and significantly higher minimum probability (0.29) compared to benign nodules (minimum 0.02). For benign nodules (n=582), Tri-Reader detected 547 (94.0%) with slightly lower median probability (0.98). The broader distribution and lower tail for benign nodules (mean $0.87 \pm 0.23$) compared to malignant (mean $0.92 \pm 0.16$) suggests that Tri-Reader's confidence scoring differentiates suspicious lesions.

### S4.2.4 Missed Nodule Characteristics

**Table S4.9** summarizes missed nodule patterns. Tri-Reader missed 41 nodules (6.3%), of which 21 (51.2%) were uniquely missed. These unique misses had median diameter 4.84 mm (range 2.1-15.8 mm), with 17/21 (81.0%) being sub-6 mm. Notably, 14 nodules were missed by both models, with mean diameter 7.59 mm. LUNA16-De unique misses (n=13) again included larger nodules (mean 14.64 mm, max 35.1 mm).

**Table S4.5. Demographics and Nodule Characteristics in the Duke Lung Cancer Screening Secret Test Cohort.**

| Category | Sub-category | Test Dataset | |
|---|---|---|---|
| **Age** | Mean (Min- Max) | 66 (54-79) | |
| **Gender** | Male | 208 (49.64%) | |
| | Female | 211 (50.36%) | |
| **Race** | White (ref) | 303 (72.32%) | |
| | Black/AA | 102 (24.34%) | |
| | Other/Unknown | 14 (3.34%) | |
| **Ethnicity** | Not Hispanic | 397 (94.75%) | |
| | Unavailable | 17 (4.06%) | |
| | Hispanic | 5 (1.19%) | |
| **Smoking status** | Currently smoking | 205 (52.70%) | |
| | Formerly smoking | 183 (47.04%) | |
| **Tobacco used years** | 30+ years | 322 (82.14%) | |
| | 21-30 years | 59 (15.05%) | |
| | 11-20 years | 9 (2.30%) | |
| | 6-10 years | 1 (0.26%) | |
| | 0-5 years | 1 (0.26%) | |
| | | **Patient** | **Nodule** |
| **Malignancy** | Benign | 378 (90.21%) | 582 (89.40%) |
| | Cancer | 41 (9.79%) | 69 (10.60%) |
| **Lung-RADS** | 1 | 2 (0.48%) | 2 (0.31%) |
| | 2 | 231 (55.13%) | 324 (49.77%) |
| | 3 | 72 (17.18%) | 129 (19.82%) |
| | 4A | 54 (12.89%) | 89 (13.67%) |
| | 4B | 39 (9.31%) | 75 (11.52%) |
| | 4X | 14 (3.34%) | 24 (3.69%) |
| | Missing | 7 (1.67%) | 8 (1.23%) |
| **Lung Cancer Type** | Adenocarcinoma | 16 (45.71%) | 22 (37.93%) |
| | Squamous cell carcinoma | 8 (22.86%) | 16 (27.59%) |
| | NSCC (other/NOS) | 3 (8.57%) | 8 (13.79%) |
| | Small cell carcinoma | 5 (14.29%) | 6 (10.34%) |
| | Other / Unknown | 3 (8.57%) | 6 (10.34%) |
| **LCS Detected Cancer** | 1 / 1.0 | 28 (82.35%) | 47 (83.93%) |
| | 0 / Unknown | 6 (17.65%) | 9 (16.07%) |
| **Nodule Size** | < 6 mm | | 413 (63.4%) |
| | 6–10 mm | | 150 (24.0%) |
| | 10–20 mm | | 71 (10.9%) |
| | > 20 mm | | 17 (0.03%) |

**Table S4.6. Size-Stratified FROC Performance (DLCS Secret Test Set, n=419 scans, 651 nodules)**

| Size Category | Model | Avg. Sensitivity | @1/8 FP | @1/4 FP | @1/2 FP | @1 FP | @2 FP | @4 FP | @8 FP | Detected/Lesions (%) | Candidates/Scans |
|---|---|---|---|---|---|---|---|---|---|---|---|
| Overall | LUNA16-De | 0.53 | 0.15 | 0.37 | 0.53 | 0.81 | 0.89 | 0.91 | 0.93 | 617/651 (94.8%) | 22.30 (9349/419) |
| | **Tri-Reader** | **0.59** | **0.21** | **0.46** | **0.60** | **0.86** | **0.92** | **0.93** | **0.95** | **610/651 (93.7%)** | **12.40 (5208/419)** |
| <6 mm (n=413) | LUNA16-De | 0.49 | 0.04 | 0.33 | 0.51 | 0.83 | 0.92 | 0.93 | 0.95 | 401/413 (97.1%) | 22.20 (6506/293) |
| | **Tri-Reader** | **0.53** | **0.10** | **0.35** | **0.53** | **0.86** | **0.93** | **0.94** | **0.95** | **389/413 (94.2%)** | **11.60 (3395/293)** |
| 6–10 mm (n=150) | LUNA16-De | 0.57 | 0.22 | 0.43 | 0.56 | 0.81 | 0.89 | 0.91 | 0.93 | 144/150 (96.0%) | 24.60 (3227/131) |
| | **Tri-Reader** | **0.61** | **0.17** | **0.52** | **0.65** | **0.84** | **0.93** | **0.94** | **0.96** | **143/150 (95.3%)** | **15.30 (2004/131)** |
| ≥10 mm (n=88) | LUNA16-De | 0.35 | 0.17 | 0.22 | 0.24 | 0.56 | 0.69 | 0.78 | 0.85 | 72/88 (81.8%) | 24.20 (1789/74) |
| | **Tri-Reader** | **0.48** | **0.15** | **0.35** | **0.44** | **0.74** | **0.84** | **0.89** | **0.92** | **78/88 (88.6%)** | **15.60 (1156/74)** |

**Table S4.7. Performance Metrics by Lung-RADS Category (DLCS Secret Test Set).** GT Count (n) denotes total ground-truth nodules; Detected (n) denotes nodules localized by the model. Mean, Median, and Min-Max summarize the detection probability distribution for each category.

| Lung-RADS | Model | GT Count (n) | Detected (n) | Probability | | |
|---|---|---|---|---|---|---|
| | | | | Mean | Median | Min–Max Range |
| 1 | LUNA16-De | 2 | 2 | 0.696 | 0.696 | 0.398-0.993 |
| | **Tri-Reader** | **2** | **2** | **0.996** | **0.996** | **0.996-0.997** |
| 2 | LUNA16-De | 324 | 321 | 0.881 | 0.984 | 0.036-1.000 |
| | **Tri-Reader** | **324** | **315** | **0.858** | **0.976** | **0.020-1.000** |
| 3 | LUNA16-De | 129 | 119 | 0.855 | 0.977 | 0.023-0.999 |
| | **Tri-Reader** | **129** | **119** | **0.878** | **0.976** | **0.066-1.000** |
| 4A | LUNA16-De | 89 | 76 | 0.822 | 0.988 | 0.037-1.000 |
| | **Tri-Reader** | **89** | **74** | **0.893** | **0.991** | **0.081-1.000** |
| 4B | LUNA16-De | 75 | 70 | 0.819 | 0.978 | 0.022-1.000 |
| | **Tri-Reader** | **75** | **70** | **0.921** | **0.987** | **0.043-1.000** |
| 4X | LUNA16-De | 24 | 21 | 0.929 | 0.995 | 0.461-1.000 |
| | **Tri-Reader** | **24** | **22** | **0.977** | **0.997** | **0.811-1.000** |

**Table S4.8. Detection Performance by Pathological Diagnosis (DLCS Secret Test Set).**

| Diagnosis | Model | GT Count (n) | Detected (n) | Probability | | |
|---|---|---|---|---|---|---|
| | | | | Mean± Std | Median | Min–Max |
| Benign | LUNA16-De | 582 | 554 | 0.85 ± 0.26 | 0.98 | 0.02-1.00 |
| | **Tri-Reader** | **582** | **547** | **0.87 ± 0.23** | **0.98** | **0.02-1.00** |
| Cancer | LUNA16-De | 69 | 63 | 0.92 ± 0.19 | 0.99 | 0.07-1.00 |
| | **Tri-Reader** | **69** | **63** | **0.92 ± 0.16** | **0.99** | **0.29-1.00** |

**Table S4.9. Missed Nodule Characteristics and Overlap Analysis (DLCS Secret Test Set).**

| Overlap Category | Count | Percentage (%) | Mean Diameter (mm) | Median Diameter (mm) |
|---|---|---|---|---|
| All Three Models | 14 | 34.1 | 7.59 ± 4.21 | 5.98 |
| LUNA16-De Only | 13 | 31.7 | 14.64 ± 8.52 | 15.31 |
| Tri-Reader Only | 21 | 51.2 | 6.55 ± 4.28 | 4.84 |

**S4.3 Lung Nodule Database (LNDbv4) Test Performances**

We used **Lung Nodule Database (LNDbv4)** [9], a chest CT dataset (2016–2018) that includes manual pulmonary nodule-level annotations and radiologist-assigned semantic characteristic ratings. Manual annotation was performed by five radiologists, with 1-3 radiologists per CT, and includes nodule segmentations and ordinal ratings (texture and related attributes). In parallel, the associated clinical radiology reports (originally in Portuguese) were curated and converted into a structured table of report-mentioned nodule entities, capturing size, location descriptors, and LNDb/LIDC-style ordinal characteristics [9, 13]. LNDbv4 additionally provides a report-to-image matching procedure that labels each record as manual-only (RadAnnotation), report-only, or matched across both sources (TextReport + RadAnnotation), using a candidate-search pipeline based on (when available) lobe/relative location, size tolerance (±3 mm), and characteristic agreement (±1 ordinal level), followed by tie-breaking using uniqueness and annotator agreement.

For our analyses, we began with 228 patients (228 CTs; 1,152 nodule-level findings) and filtered to a report-linkage-eligible subset of 212 patients (212 CTs; 768 findings). For analyses requiring text-image concordance, we retained only matched findings (TextReport + RadAnnotation) to ensure that each analyzed nodule had both spatial ground truth and corresponding report-derived descriptors, and we excluded RadAnnotation-only (manual-only) findings without a report entity. Finally, for detection-versus-missed evaluation, we used a consensus evaluation cohort of 155 patients (155 CTs; 377 nodules). In the consensus evaluation cohort (155 CTs; 377 nodules), lesion type was dominated by small findings: micronodules accounted for 183/377 (48.5%) and nodules for 175/377 (46.4%), whereas granulomas were infrequent (17/377, 4.5%) and masses were rare (2/377, 0.5%). Figure S4.1 shown the inclusion and exclusion criteria used in this study.

**Nodule size** from manual spatial annotations was $6.11 \pm 4.06$ mm (median 4.97 mm, IQR 3.90-6.52 mm; range 3-30 mm). On ordinal semantic scales, nodules were predominantly solid-leaning (Texture $4.48 \pm 0.93$, median 5, IQR 4-5) with calcification typically absent/minimal (Calcification $5.61 \pm 0.85$, median 6, IQR 6–6), and internal structure largely consistent with soft tissue (Internal structure $1.03 \pm 0.21$, median 1, IQR 1-1). Morphologic irregularity was generally limited (Lobulation $1.55 \pm 0.72$; Spiculation $1.32 \pm 0.65$), with relatively well-defined borders (Margin $4.21 \pm 0.88$) and moderately round shape (Sphericity $3.72 \pm 1.08$). Overall malignancy suspicion was low-to-intermediate (Malignancy $2.40 \pm 1.02$), and conspicuity was moderate-to-high (Subtlety $3.92 \pm 1.09$, median 4, IQR 3–5).

**Radiologist agreement** was summarized using canonical vote-pattern categories, where 1R denotes a single-reader annotation (1/1 positive vote), 2R denotes two-reader review (2/2 consensus or 1/2 disagreement), and 3R denotes three-reader review (3/3 consensus or 2/3 with 1 dissent); No votes indicates missing vote information and Other captures non-standard vote patterns. Across all 377 nodules, most were single-reader (1R (1/1): 206/377, 54.6%), followed by two-reader consensus (2R (2/2) Consensus: 106/377, 28.1%) and three-reader consensus (3R (3/3) Consensus: 32/377, 8.5%), with fewer disagreement patterns (2R (1/2) Disagree: 24/377, 6.4%; 3R (2/3) 1 dissent: 9/377, 2.4%).

### S4.3.1 Overall and Size-Stratified Detection Performance

**Table S4.10** shown FROC results for LNDbv4. Tri-Reader achieved CPM 0.47 (95% CI: 0.42-0.52) with 11.34 candidates/scan, the lowest candidate burden while maintaining performance equivalent to LUNA17-De. Size-stratified analysis showed strong performance for actionable nodules. For nodules ≥10 mm (n=38), Tri-Reader achieved average sensitivity 0.70 with 100% detection at 8 FP/scan, identifying all 37 detected nodules (one nodule missed by all models). For 6<10 mm nodules (n=74), average sensitivity was 0.60 with only 14.37 candidates/scan. For 6 mm micronodules (n=265), average sensitivity was 0.31, reflecting the inherent challenge of detecting subtle, small lesions. 0.51 (0.46-0.56)

### S4.3.2 Detection by Radiologist Consensus Pattern

**Table S4.11** presents detection performance stratified by radiologist agreement patterns. Model detection probability showed strong correlation with consensus: for three-reader unanimous agreement (3R consensus, n=32), both models assigned high mean probabilities (Tri-Reader: 0.94 ± 0.13 and LUNA16-De: 0.95 ± 0.19). **For two-reader consensus** (2R consensus, n=106), Tri-Reader maintained performance (mean probability 0.84 ± 0.24) with tight distributions (IQR: 0.71-0.98). Importantly, for single-reader annotations (1R, n=206), Tri-Reader showed greater dispersion (0.77 ± 0.32) with more low-probability outliers, indicating appropriate calibration to nodule conspicuity. **For disagreement cases** (2R disagree, n=24), Tri-Reader demonstrated notably lower and more dispersed confidence (mean 0.73 ± 0.32, minimum 0.07) compared to LUNA16-De, suggesting that Tri-Reader's confidence scoring aligns with inter-observer variability: nodules with low radiologist agreement receive appropriately lower detection confidence.

### S4.3.3 Semantic Characteristics Analysis

**Table S4.12** shown pooled analysis (across both the models) comparing semantic characteristics of detected vs missed nodules. Subtlety showed the largest separation (mean detected 4.06 vs missed 3.15; Cohen's d = 0.88; $p < 0.001$)[10], confirming that conspicuity is the primary driver of detectability. Malignancy rating (d = 0.66; $p < 0.001$) and **diameter** (d = 0.60; $p < 0.001$) showed medium effect sizes. Lobulation (d = 0.53) and Spiculation (d = 0.42) demonstrated moderate associations with detection, indicating that morphologically suspicious features improve detectability. Margin showed no significant association (p = 0.081), while Sphericity showed a small negative association (d = -0.31; $p < 0.001$), suggesting that more irregular shapes are easier to detect. **Table S4.13** ranks detectability drivers by model-specific Cohen's d values [10]. Notably, Tri-Reader demonstrated higher sensitivity to **Spiculation** (d = 0.51) compared to LUNA16-De (d = 0.39), suggesting that Tri-Reader preferentially detects nodules with suspicious morphological features associated with malignancy.

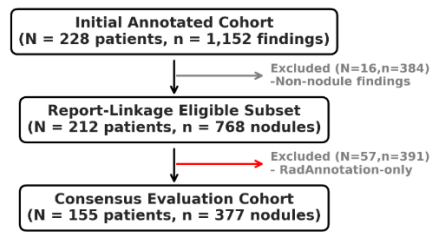

**Figure S4.1.** Cohort construction and inclusion/exclusion criteria for LNDbv4, including the image-annotated cohort, report-linkage-eligible subset, matched-only (TextReport + RadAnnotation) analytic cohort, and the consensus evaluation cohort used for detection-versus-missed analyses.

**Table S4.10. Detection Sensitivity (95% CI) Across Average False Positives per Scan, Stratified by Nodule Size (LNDbv4).**

| Size bin | Model | CPM | @1/8 | @1/4 | @1/2 | @1 | @2 | @4 | @8 | Detected/lesions (%) | Candidate/Scans |
|---|---|---|---|---|---|---|---|---|---|---|---|
| Overall | LUNA16-De | 0.51 (0.46-0.56) | 0.18 (0.13-0.26) | 0.31 (0.21-0.39) | 0.42 (0.36-0.49) | 0.52 (0.46-0.58) | 0.62 (0.56-0.69) | 0.72 (0.66-0.77) | 0.80 (0.75-0.85) | 321/377 (85.1%) | 19.30 (2992/155) |
| Overall | Tri-Reader | 0.47 (0.42-0.52) | 0.19 (0.13-0.24) | 0.24 (0.19-0.30) | 0.35 (0.27-0.41) | 0.47 (0.41-0.54) | 0.55 (0.49-0.61) | 0.69 (0.62-0.75) | 0.78 (0.72-0.84) | 297/377 (78.8%) | 11.34 (1757/155) |
| <6 | LUNA16-De | 0.35 (0.30-0.39) | 0.00 (0.00-0.01) | 0.04 (0.00-0.09) | 0.18 (0.09-0.27) | 0.35 (0.28-0.43) | 0.49 (0.42-0.56) | 0.62 (0.56-0.70) | 0.74 (0.68-0.80) | 216/265 (81.5%) | 19.77 (2313/117) |
| <6 | Tri-Reader | 0.31 (0.27-0.36) | 0.02 (0.00-0.07) | 0.07 (0.03-0.13) | 0.14 (0.09-0.20) | 0.26 (0.19-0.33) | 0.42 (0.35-0.49) | 0.55 (0.48-0.64) | 0.70 (0.63-0.77) | 189/265 (71.3%) | 11.44 (1339/117) |
| 6 to <10 | LUNA16-De | 0.60 (0.50-0.68) | 0.22 (0.04-0.33) | 0.28 (0.17-0.49) | 0.50 (0.28-0.67) | 0.65 (0.53-0.77) | 0.78 (0.63-0.88) | 0.85 (0.77-0.92) | 0.89 (0.83-0.95) | 68/74 (91.9%) | 21.83 (1179/54) |
| 6 to <10 | Tri-Reader | 0.60 (0.52-0.69) | 0.30 (0.16-0.42) | 0.32 (0.23-0.46) | 0.41 (0.29-0.58) | 0.62 (0.51-0.78) | 0.77 (0.65-0.87) | 0.85 (0.74-0.94) | 0.95 (0.86-0.99) | 73/74 (98.6%) | 14.37 (776/54) |
| >=10 | LUNA16-De | 0.77 (0.63-0.87) | 0.63 (0.19-0.76) | 0.63 (0.48-0.78) | 0.66 (0.50-0.82) | 0.71 (0.57-0.88) | 0.87 (0.71-0.97) | 0.92 (0.81-1.00) | 0.95 (0.86-1.00) | 37/38 (97.4%) | 15.16 (470/31) |
| >=10 | Tri-Reader | 0.70 (0.56-0.84) | 0.18 (0.03-0.63) | 0.45 (0.11-0.73) | 0.63 (0.44-0.82) | 0.79 (0.57-0.92) | 0.87 (0.75-0.97) | 0.92 (0.82-1.00) | 1.00 (0.86-1.00) | 35/38 (92.1%) | 9.06 (281/31) |

**Table S4.11. Detection probability stratified by radiologist consensus pattern in LNDbv4.**

| Radiologist Consensus Pattern | Model | GT (n) | Detected (n) | Detection Probability (Mean ± SD) | Range (Min–Max) |
|---|---|---|---|---|---|
| 1R (1/1) | LUNA16-De | 206 | 155 | 0.76 ± 0.34 | 0.02-1.00 |
| 1R (1/1) | Tri-Reader | 206 | 136 | 0.77 ± 0.32 | 0.02-1.00 |
| 2R (2/2) Consensus | LUNA16-De | 106 | 103 | 0.91 ± 0.19 | 0.16-1.00 |
| 2R (2/2) Consensus | Tri-Reader | 106 | 99 | 0.84 ± 0.24 | 0.05-1.00 |
| 2R (1/2) Disagree | LUNA16-De | 24 | 23 | 0.88 ± 0.19 | 0.38-1.00 |
| 2R (1/2) Disagree | Tri-Reader | 24 | 24 | 0.73 ± 0.32 | 0.07-1.00 |
| 3R (3/3) Consensus | LUNA16-De | 32 | 31 | 0.95 ± 0.19 | 0.09-1.00 |
| 3R (3/3) Consensus | Tri-Reader | 32 | 30 | 0.94 ± 0.13 | 0.35-1.00 |
| 3R (2/3) 1 dissent | LUNA16-De | 9 | 9 | 0.96 ± 0.05 | 0.88-1.00 |
| 3R (2/3) 1 dissent | Tri-Reader | 9 | 8 | 0.81 ± 0.30 | 0.18-1.00 |

**Table S4.12. Semantic Characteristics: Detected vs Missed Nodules (Pooled Analysis, LNDbv4)**

| Characteristic | Detected Mean | Missed Mean | Mean Difference | Mann-Whitney p | Cohen's d | Effect Size |
|---|---|---|---|---|---|---|
| DiamEq_Rad (mm) | 6.472 | 4.109 | 2.363 | <0.001* | 0.60 | Medium |
| Texture (1–5) | 4.451 | 4.619 | −0.168 | 0.003* | −0.18 | Small |
| Malignancy (0–5) | 2.497 | 1.847 | 0.650 | <0.001* | 0.66 | Medium |
| Subtlety (1–5) | 4.058 | 3.148 | 0.909 | <0.001* | 0.88 | Large |
| Spiculation (1–5) | 1.364 | 1.094 | 0.270 | <0.001* | 0.42 | Small |
| Lobulation (1–4) | 1.603 | 1.227 | 0.376 | <0.001* | 0.53 | Medium |
| Margin (1–5) | 4.191 | 4.300 | −0.109 | 0.081 | −0.12 | Small |
| Sphericity (1–5) | 3.673 | 4.008 | −0.335 | <0.001* | −0.31 | Small |

**Note:** Pooled analysis includes 957 detected and 174 missed lesion-model pairs across both models. Subtlety (conspicuity) shows the largest effect size, followed by Malignancy rating and Diameter. **Asterisk:** indicates significance after Bonferroni correction ($\alpha^* = 0.00625$ for 8 tests). Higher Subtlety scores indicate more conspicuous nodules (easier to see) *.

**Table S4.13. Top Detectability Factors by Model (LNDbv4).** DiamEq_Rad = diameter measured by radiologist.

| Rank | LUNA16-De | Cohen's d | Tri-Reader | Cohen's d |
|---|---|---|---|---|
| 1 | Subtlety | 0.94 | Subtlety | 1.00 |
| 2 | Malignancy | 0.60 | Malignancy | 0.71 |
| 3 | DiamEq_Rad | 0.57 | DiamEq_Rad | 0.66 |
| 4 | Lobulation | 0.50 | Lobulation | 0.58 |
| 5 | Spiculation | 0.39 | Spiculation | 0.51 |

## S4.4 Integrated Multiomics (IMD-CT) Dataset Test Performances

The external IMD-CT test cohort included 2,032 patients/CT scans, each with one annotated indeterminate pulmonary nodule, as described by Zhao et al. [14]. The mean age was 55.9 ± 12.3 years, with 434/2032 (21.4%) benign and 1598/2032 (78.6%) malignant nodules. Mean solid component size was 15.7 ± 13.2 mm, and mean nodule diameter was 30.0 ± 20.8 mm. There were no 6 mm nodules; most nodules were 10–20 mm (37.9%), 20-30 mm (26.8%), or >30 mm (33.8%) (**Table S4.14**).

### S4.3.1 Overall and Size-Stratified Detection Performance

**Table S4.15** shown the FROC results for the overall and size-stratified subsets for the IMD-CT cohort. Tri-Reader achieved the CPM (0.73; 95% CI: 0.72-0.75), compared to LUNA16-De (0.63; 0.62-0.65). Importantly, Tri-Reader maintaining the lowest candidate burden, 6.71 candidates/scan (13,642/2,032) versus 13.91 candidates/scan (28,258/2,032) for LUNA16-De, representing a ~2× reduction compared to LUNA16-De. At clinically relevant operating points (1-2 FP/scan), Tri-Reader maintained sensitivities of 0.80-0.87, substantially higher than baseline models. For the $10 \leq 20$ mm nodule subset (n=771, 37.9% of cohort), Tri-Reader achieved average sensitivity 0.76 with only 4.89 candidates/scan. For nodules ≥20 mm (n=1,231, 60.6%), average sensitivity was 0.73 with 7.89 candidates/scan.

**Table S4.14. Summary of patient, CT scan, and lesion characteristics.** Values are shown as count (percentage) for categorical variables and mean ± standard deviation for continuous variables.

| Category | Value | Total (n = 2032) |
|---|---|---|
| Patient | Unique Patients | 2032 (100.0%) |
| Age (years) | Mean ± SD | 55.9 ± 12.3 |
| CT Scans | Unique CTs | 2032 (100.0%) |
| Nodule | Unique Annotations | 2032 (100.0%) |
| Cancer | Benign (0) | 434 (21.4%) |
|  | Malignant (1) | 1598 (78.6%) |
| Solid Size (mm) | Mean ± SD | 15.7 ± 13.2 |
| Diameter (mm) | Mean ± SD | 30.0 ± 20.8 |
| Nodule Size | Sub centimeter (<6 mm) | 0 (0.0%) |
|  | Small (6-10 mm) | 30 (1.5%) |
|  | Intermediate (10-20 mm) | 771 (37.9%) |
|  | Large (20-30 mm) | 545 (26.8%) |
|  | Very Large (>30 mm) | 686 (33.8%) |

**Table S4.15. IMD-CT cohort characteristics. Patient, CT, and nodule-level characteristics for the IMD-CT test cohort (n = 2,032).** Categorical variables are reported as count (percentage); continuous variables are reported as mean ± standard deviation.

|  | Model | Sensitivity (95% CI) @ Avg. FPs/Scans | | | | | | | Detected /lesions (%) | Candidate/Scans |
|---|---|---|---|---|---|---|---|---|---|---|
|  |  | Avg. | @1/8 | @1/4 | @1/2 | @1 | @2 | @4 | @8 | | |
| Overall | LUNA 16-De | 0.63 (0.62-0.65) | 0.27 (0.25-0.30) | 0.39 (0.37-0.42) | 0.54 (0.51-0.56) | 0.66 (0.64-0.69) | 0.80 (0.78-0.82) | 0.87 (0.85-0.88) | 0.91 (0.89-0.92) | 1874/2032 (92.2%) | 13.91 (28258/2032) |
|  | Tri-Reader | 0.73 (0.72-0.75) | 0.35 (0.32-0.39) | 0.52 (0.49-0.55) | 0.69 (0.66-0.71) | 0.80 (0.79-0.83) | 0.87 (0.85-0.88) | 0.90 (0.89-0.91) | 1.00 (1-1) | 1870/2032 (92.0%) | 6.71 (13642/2032) |
| **Size bin** | | | | | | | | | | | |
| <10 mm | LUNA 16-De | 0.50 (0.39-0.62) | 0.03 (0.00-0.23) | 0.17 (0.00-0.33) | 0.37 (0.13-0.57) | 0.63 (0.43-0.80) | 0.77 (0.60-0.90) | 0.80 (0.67-0.93) | 0.80 (0.67-0.93) | 24/30 (80.0%) | 10.87 (326/30) |
|  | Tri-Reader | 0.54 (0.43-0.66) | 0.03 (0.00-0.27) | 0.17 (0.00-0.40) | 0.40 (0.13-0.63) | 0.63 (0.43-0.77) | 0.73 (0.50-0.90) | 0.80 (0.63-1.00) | 1.00 (1.00-1.00) | 24/30 (80.0%) | 5.17 (155/30) |
| 10-<20 mm | LUNA 16-De | 0.71 (0.68-0.74) | 0.36 (0.30-0.42) | 0.52 (0.47-0.57) | 0.67 (0.62-0.70) | 0.78 (0.74-0.81) | 0.86 (0.83-0.88) | 0.89 (0.87-0.91) | 0.92 (0.90-0.94) | 714/771 (92.6%) | 12.30 (9487/771) |
|  | **Tri-Reader** | **0.76 (0.73-0.79)** | **0.38 (0.33-0.44)** | **0.55 (0.50-0.61)** | **0.72 (0.67-0.76)** | **0.82 (0.79-0.85)** | **0.88 (0.86-0.91)** | **1.00 (0.90-1)** | **1.00 (1-1)** | 708/771 (91.8%) | 4.89 (3769/771) |
| ≥20 mm | LUNA 16-De | 0.59 (0.57-0.61) | 0.26 (0.22-0.29) | 0.34 (0.31-0.37) | 0.46 (0.42-0.49) | 0.59 (0.56-0.63) | 0.75 (0.72-0.78) | 0.85 (0.83-0.87) | 0.90 (0.88-0.91) | 1136/1231 (92.3%) | 14.98 (18445/1231) |
|  | **Tri-Reader** | **0.73 (0.71-0.75)** | **0.34 (0.29-0.38)** | **0.50 (0.47-0.54)** | **0.68 (0.64-0.71)** | **0.81 (0.78-0.83)** | **0.87 (0.86-0.89)** | **0.90 (0.88-0.92)** | **1.00 (1-1)** | 1138/1231 (92.4%) | 7.89 (9718/1231) |

**Reference**

[1] F. I. Tushar *et al.*, "AI in Lung Health: Benchmarking Detection and Diagnostic Models Across Multiple CT Scan Datasets," *arXiv preprint arXiv:2405.04605,* 2024.
[2] A. A. A. Setio *et al.*, "Validation, comparison, and combination of algorithms for automatic detection of pulmonary nodules in computed tomography images: The LUNA16 challenge," *Med Image Anal,* vol. 42, pp. 1-13, Dec 2017, doi: 10.1016/j.media.2017.06.015.
[3] D. Peeters, B. Obreja, N. Antonissen, and C. Jacobs, "The LUNA25 Challenge: Public Training and Development set - Imaging Data," doi: 10.5281/zenodo.14223624.
[4] F. I. Tushar *et al.*, "Virtual Lung Screening Trial (VLST): An In Silico Study Inspired by the National Lung Screening Trial for Lung Cancer Detection," *Medical Image Analysis,* p. 103576, 2025.
[5] F. I. Tushar *et al.*, "SYN-LUNGS: Towards Simulating Lung Nodules with Anatomy-Informed Digital Twins for AI Training," *arXiv preprint arXiv:2502.21187,* 2025.
[6] Y. He *et al.*, "VISTA3D: A unified segmentation foundation model for 3D medical imaging," in *Proceedings of the Computer Vision and Pattern Recognition Conference*, 2025, pp. 20863-20873.
[7] J. Wasserthal *et al.*, "TotalSegmentator: robust segmentation of 104 anatomic structures in CT images," *Radiology: Artificial Intelligence,* vol. 5, no. 5, p. e230024, 2023.
[8] F. Isensee, P. F. Jaeger, S. A. Kohl, J. Petersen, and K. H. Maier-Hein, "nnU-Net: a self-configuring method for deep learning-based biomedical image segmentation," *Nature methods,* vol. 18, no. 2, pp. 203-211, 2021.
[9] C. A. Ferreira *et al.*, "LNDb v4: pulmonary nodule annotation from medical reports," *Sci Data,* vol. 11, no. 1, p. 512, May 17 2024, doi: 10.1038/s41597-024-03345-6.
[10] J. Cohen, *Statistical power analysis for the behavioral sciences*. routledge, 2013.
[11] H. B. Mann and D. R. Whitney, "On a test of whether one of two random variables is stochastically larger than the other," *The annals of mathematical statistics,* pp. 50-60, 1947.
[12] A. J. Wang *et al.*, "The Duke Lung Cancer Screening (DLCS) Dataset: A Reference Dataset of Annotated Low-Dose Screening Thoracic CT," *Radiol Artif Intell,* vol. 7, no. 4, p. e240248, Jul 2025, doi: 10.1148/ryai.240248.
[13] S. G. Armato III *et al.*, "The lung image database consortium (LIDC) and image database resource initiative (IDRI): a completed reference database of lung nodules on CT scans," *Medical physics,* vol. 38, no. 2, pp. 915-931, 2011.
[14] M. Zhao *et al.*, "Integrated multiomics signatures to optimize the accurate diagnosis of lung cancer," *Nat Commun,* vol. 16, no. 1, p. 84, Jan 2 2025, doi: 10.1038/s41467-024-55594-z.